\documentclass[letterpaper]{article}
\usepackage[preprint]{aaai2027}

\usepackage[hyphens]{url}
\usepackage{graphicx}
\usepackage{natbib}
\usepackage{caption}
\usepackage{amsmath}
\usepackage{amsfonts}
\usepackage{amssymb}
\usepackage{booktabs}
\usepackage{subcaption}
\usepackage{multirow}

\title{Revisiting Shape and Texture Reliance with Category-Separability-Calibrated Suppression}
\author{
    Ning Jiang\textsuperscript{1,2,*},
    Tianyi Luo\textsuperscript{4,*},
    Zhengyong Huang\textsuperscript{1,2},
    Yao Sui\textsuperscript{1,2,3,$\dagger$}
}

\affiliations{
    \textsuperscript{1}Institute of Medical Technology, Peking University Health Science Center, Beijing 100191, China\\
    \textsuperscript{2}National Institute of Health Data Science, Peking University, Beijing 100191, China\\
    \textsuperscript{3}Institute for Artificial Intelligence, Peking University, Beijing 100871, China\\
    \textsuperscript{4}School of Computer Science and Engineering, Sun Yat-sen University, Guangzhou, China\\
    \textsuperscript{*}These authors contributed equally.\\
    \textsuperscript{$\dagger$}Corresponding author: Yao Sui
    (\texttt{yaosui@pku.edu.cn})
}

\begin{document}

\maketitle

\begin{abstract}
Feature-suppression evaluations infer model reliance on shape or texture from the accuracy loss caused by attenuating each type of information. Such losses, however, conflate feature reliance with the amount of category-relevant information removed by the corresponding transformation. Because shape and texture are suppressed using different operators, their effects are not directly comparable. We introduce the Semantic Degradation Index (SDI), which quantifies the suppression-induced reduction in category separability relative to clean images in a fixed clean-reference discriminative space constructed from handcrafted features. On an ImageNet16-like benchmark, we use SDI to compare Gaussian blur for texture suppression with grid distortion for shape suppression over their overlapping degradation range. At comparable SDI values, all five evaluated ImageNet-trained convolutional neural networks (CNNs) retain less accuracy under Gaussian blur than under grid distortion. This results supports stronger texture than shape reliance under the evaluated operators, contrasting with the shape-dominant conclusion obtained from unmatched suppression conditions. The evaluated Vision Transformers (ViTs) also generally retain more accuracy than CNNs under both operators. To determine whether this advantage extends beyond classification, We evaluate fixed brain-encoding models using clean and suppressed images from the Natural Scenes Dataset. Under both operators, ViT features show smaller suppression-induced decreases in noise-ceiling-normalized explained variance than CNN features. These findings establish category separability as an important reference for interpreting suppression-based feature reliance and show that the CNN-ViT robustness difference extends to model representations predictive of human visual cortical responses.
\end{abstract}


\section{Introduction}

Understanding which visual features support a model's predictions is central to interpreting visual recognition systems. Shape and texture have received particular attention because two common evaluation paradigms appear to offer a different view of their importance. In cue-conflict images, where the shape of one category is combined with the texture of another, ImageNet-trained convolutional neural networks (CNNs) often follow texture, whereas humans more often follow shape~\citep{geirhos2018imagenet}. Cue-conflict experiments reveal \emph{cue preference}: which cue dominates when cues disagree. They do not directly reveal \emph{feature reliance}: how much successful recognition depends on each cue when classifying ordinary images.

Suppression-based evaluation addresses feature reliance by attenuating one type of information and measuring the resulting loss in classification accuracy. Under this paradigm, prior work~\citep{burgert2026imagenet} reported larger accuracy drops after shape suppression than after texture suppression for ImageNet-trained CNNs, suggesting stronger reliance on shape. The inference, however, assumes that the suppression conditions being compared constitute interventions of comparable strength. An accuracy drop reflects not only how much a model relies on the targeted feature, but also how much category-relevant information the transformation removes. A larger drop can therefore indicate stronger feature reliance, greater transformation-induced damage, or both.

\begin{figure}[htbp]
    \centering
    \includegraphics[width=0.9\linewidth]{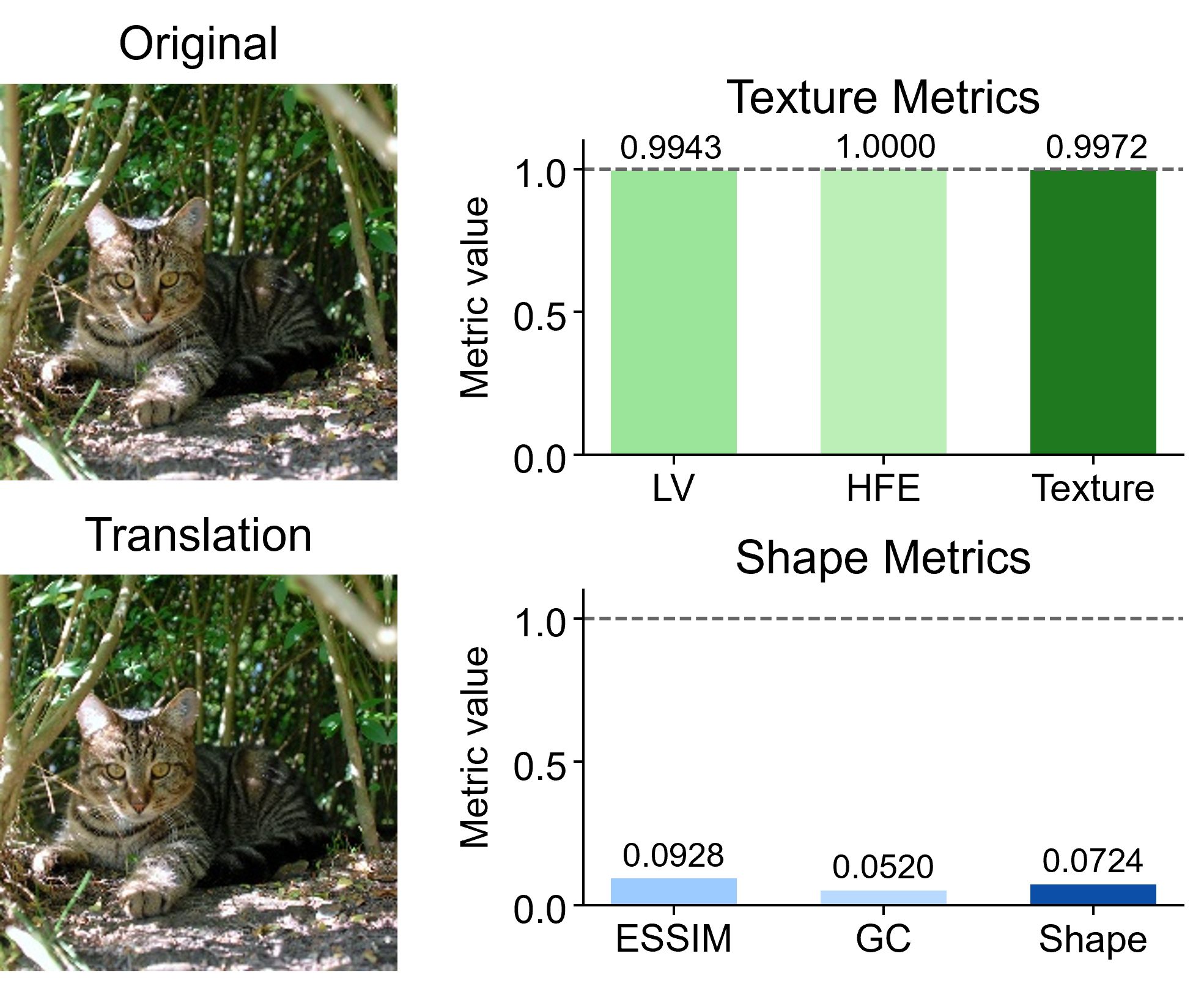}
    \caption{A five-pixel horizontal translation preserves category but substantially reduces the alignment-sensitive shape metrics, illustrating that low-level metric changes need not correspond to category-level damage. The rightmost bar in each panel averages the preceding two metrics (Texture: LV and HFE; Shape: ESSIM and GC). Values closer to 1 indicate greater preservation relative to the clean image.}
    \label{fig:cat-shifting-5-pixels}
\end{figure}

The comparability problem is especially acute across feature types because shape and texture are suppressed by different transformations with unrelated parameterizations. A blur standard deviation and a spatial-distortion magnitude, for example, cannot be directly equated. Feature-specific metrics do not fully resolve this problem. Local variance (LV) and high-frequency energy (HFE) can verify texture attenuation, while edge-structure similarity (ESSIM) and gradient correlation (GC) can verify shape disruption~\citep{burgert2026imagenet}; however, these quantities measure different low-level image properties and do not define a common scale for category-level damage. A small image translation illustrates the distinction: it can sharply reduce alignment-sensitive shape scores while leaving category identity intact (Fig.~\ref{fig:cat-shifting-5-pixels}). Thus, validating that each operator affects its intended property is necessary, but insufficient for comparing the resulting classification losses.

We address this confound by comparing suppression conditions at similar estimated losses of category separability. Specifically, we introduce the Semantic Degradation Index (SDI). Clean and suppressed images are represented using the same fixed handcrafted features, and a discriminative space is learned from clean images only. SDI measures the reduction, relative to this clean reference, in the ratio of between-category to within-category scatter. Because every suppression condition is projected into the same fixed space, SDI does not replace feature-specific validation: the latter tests whether an operator targets the intended image property, whereas SDI estimates how much category separability is lost as a consequence.

We apply this framework to an ImageNet16-like benchmark~\citep{geirhos2018imagenet,geirhos2018generalisation}. A pilot analysis evaluates color, shape, and texture suppression and motivates the main focus on the latter two, for which suppression strength can be varied over broad parameter grids. We use Gaussian blur to suppress texture grid distortion to disrupt shape, validate their intended low-level effects, and compare model performance over the SDI range shared by the two operators. This design lets us ask whether the conclusion about shape-texture reliance changes once suppression-induced category damage is placed on a common scale.

We further ask whether robustness under feature suppression differs between CNNs and Vision Transformers (ViTs), and whether any such difference extends beyond classification. Classification robustness alone does not establish that the preserved representations remain informative about human vision. We therefore complement the behavioral analysis with neural encoding \citep{naselaris2011encoding,yamins2014performance}: encoding models fitted on clean-image feature predict visual cortical responses from the Natural Scenes Dataset (NSD)~\citep{allen2022massive}, and the same fixed models are evaluated using features from clean and suppressed images. The resulting decline measures how strongly suppression disrupts model features that account for human visual cortical responses.

Our contributions are threefold. First, we identify unequal category-level damage as a confound in cross-feature suppression comparisons and introduce SDI as a common clean-referenced scale for reducing this confound. Second, at comparable SDI values, Gaussian blur yields lower retained accuracy than grid distortion for every evaluated CNN, supporting stronger texture than shape reliance under the evaluated operators and revising the conclusion obtained from unmatched suppression conditions. Third, the evaluated ViTs generally retain more classification accuracy than CNNs under both operators and show smaller suppression-induced declines in cortical-response prediction. These results establish a more controlled basis for interpreting suppression-induced accuracy drops and show that the CNN-ViT robustness difference extends from classification behavior to representations predictive of human visual cortical activity.

\section{Related Work}

\subsection{From cue preference to feature reliance}

Cue-conflict evaluation has become a standard approach for studying the relative influence of shape and texture on visual recognition. Geirhos et al.~\citep{geirhos2018imagenet} combined the shape of one category with the texture of another and observed that ImageNet-trained CNNs more often predicted the texture category, whereas humans more often followed shape. Because the cues support incompatible labels, these stimuli directly reveal which cue dominates a model's final decision.

The resulting preference, however, depends on both the model and the evaluation design. Training data and augmentation can substantially change measured shape-texture preference~\citep{hermann2020origins}, while cue validity, recognizability, cue balance, and the scoring label space can alter the resulting estimate~\citep{kim2026reliability}. Moreover, a texture-dominated prediction does not imply that shape information is absent: shape may remain encoded in intermediate representations even when it does not determine the final output~\citep{hermann2020origins,islam2021shape}. Cue-conflict evaluation therefore measures cue dominance under disagreement, rather than how much ordinary recognition depends on each feature. The latter question motivates feature-reliance evaluation.

\subsection{Feature suppression and cross-operator comparability}

Feature reliance is commonly studied through interventions that remove or attenuate selected visual information and measure the resulting change in recognition. Prior work has manipulated contours, spatial arrangements, and frequency content to identify the information supporting CNN classification~\citep{jo2017measuring,baker2018deep,wang2020high,farahat2023novel}. Such studies provide direct evidence that performance depends on the modified information, but most consider a single visual property and therefore do not require effects from different operators to be placed on a common scale.

Burgert et al.~\citep{burgert2026imagenet} brought shape, texture, and color interventions into one feature-reliance framework. Under their evaluated conditions, shape suppression caused a larger accuracy loss than texture suppression for ImageNet-trained CNNs, suggesting stronger shape reliance. This results differs from the texture preference typically observed in cue-conflict evaluation, but the two findings are not inherently contradictory: cue conflict measures which feature wins when cues disagree, whereas suppression measures how performance changes when information is removed.

A separate comparability issue arises within suppression-based evaluation. Different feature types require difficult operators, whose parameters and image effects are not commensurate. Consequently, a cross-operator difference in accuracy loss can reflect both model reliance and unequal destruction of category-relevant information. Feature-specific metrics can validate that each operator changes its intended low-level property, but cannot determine whether two operators preserve comparable amounts of category structure. Our work addresses this missing comparison scale by measuring every suppression condition in the same clean-reference discriminative space and comparing shape and texture at similar reductions in category separability.




\subsection{CNN-ViT robustness from classification to neural encoding}

CNNs and Vision Transformers (ViTs) have been compared under occlusion, image perturbations, common corruption, and distribution shifts~\cite{bhojanapalli2021understanding, paul2022vision, zhang2022delving,naseer2021intriguing,bai2021transformers}. Several studies found higher robustness for the evaluated ViTs, although the relative advantage can depend strongly on architecture and training recipe~\citep{wang2023cnnsrobust}. These results motivate an explicitly scoped comparison rather than a universal claim about either model family. We extend this literature by comparing the evaluated CNNs and ViTs under controlled shape and texture suppression, using identical transformed inputs and evaluation criteria within each suppression family. 

Robust classification does not by itself establish that a model preserves representations relevant to human vision. human-model comparisons in shape-texture research have largely focused on behavioral outputs, particularly cue-conflict decisions. Neural encoding provides a complementary representational test by learning mappings from model features to stimulus-evoked cortical responses~\citep{naselaris2011encoding,yamins2014performance,schrimpf2018brainscore,kubilius2019brainlike}. The Natural Scenes Dataset (NSD), which contains high-resolution 7T fMRI measurements from eight participants viewing thousands of natural images, enables such comparisons across visual cortex~\citep{allen2022massive}. We use fixed encoding models to ask whether the CNN-ViT difference under feature suppression extends from retained classification accuracy to the preservation of feature predictive of visual cortical responses.

\section{Methods}

\subsection{ImageNet16-like benchmark and suppression design}

An ImageNet16-like benchmark was constructed following the dataset protocol of Burgert et al.~\citep{burgert2026imagenet}. The benchmark contains 16 entry-level categories derived from the WordNet hierarchy~\citep{miller1995wordnet}, with 50 images per category selected from the ImageNet validation set using their subclass-balanced sampling protocol. The resulting dataset contains 800 images, all resized to $224 \times 224$ pixels.

Three transformations are used to suppress color, shape, and texture information. Color suppression is implemented by grayscale conversion and evaluated as a single fixed condition. Shape and texture suppression are varied over multiple strength levels.

Texture suppression is implemented using Gaussian blur, with suppression strength controlled by the kernel size $k$ and standard deviation $\sigma$. Shape suppression is implemented using grid distortion, replacing the patch-shuffle operation used in prior work. Patch shuffle rearranges discrete image regions and introduces discontinuities at patch boundaries. Grid distortion instead applies interpolated spatial displacements over a regular control grid, with suppression strength controlled by the grid density and the maximum displacement magnitude.

Gaussian blur and grid distortion are evaluated over broad parameter grids, with all other implementation settings held fixed. The ranges and sampling intervals of the varied parameters are detailed in the supplementary material. Their effects on category separability are quantified using the Semantic Degradation Index (SDI) described in the next subsection.

\begin{figure}[htbp]
    \centering
    \includegraphics[width=\linewidth]{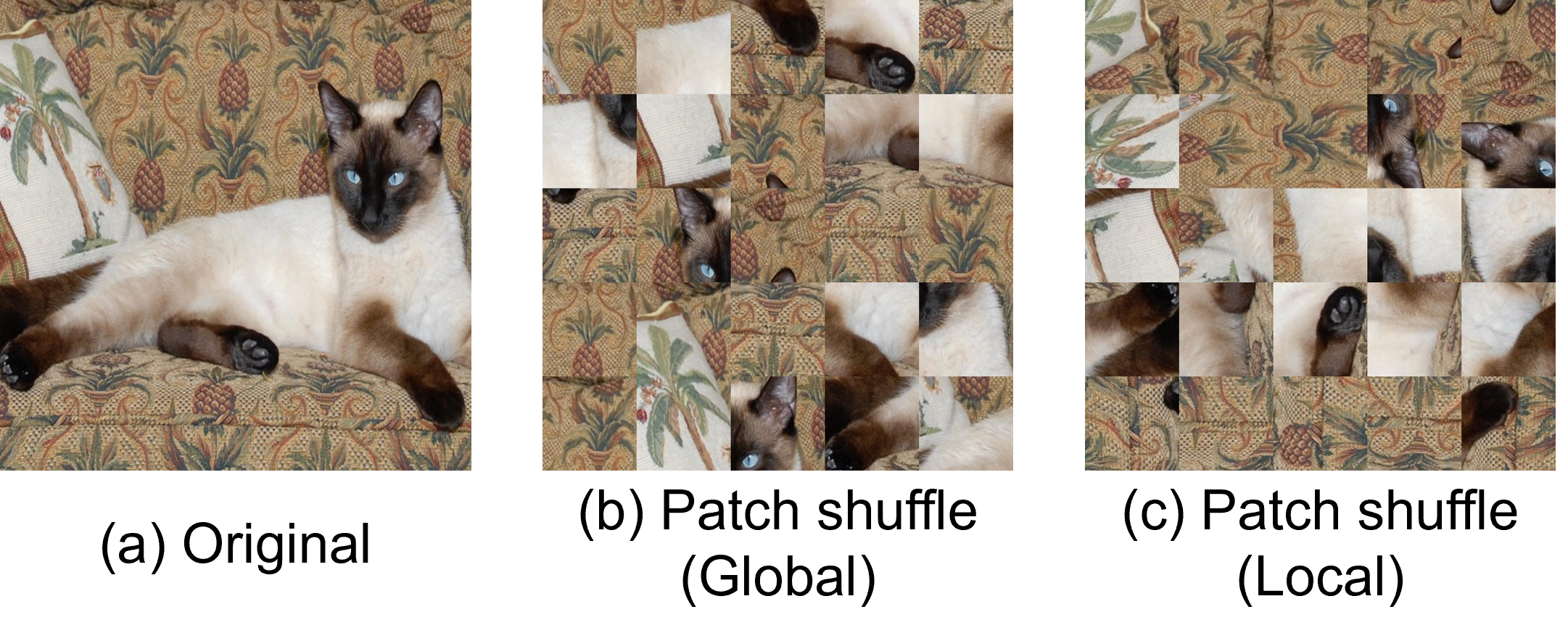}
    \caption{Examples of patch-shuffle-based shape suppression. Rearranging image regions changes their spatial organization, can recombine foreground and background content, and introduces discontinuities at patch boundaries. These effects motivate the use of grid distortion as the shape-suppression operator in our experiments.}
    \label{fig:patch-based-rearrangement}
\end{figure}

\begin{figure*}[t]
    \centering
    \includegraphics[width=0.8\textwidth]{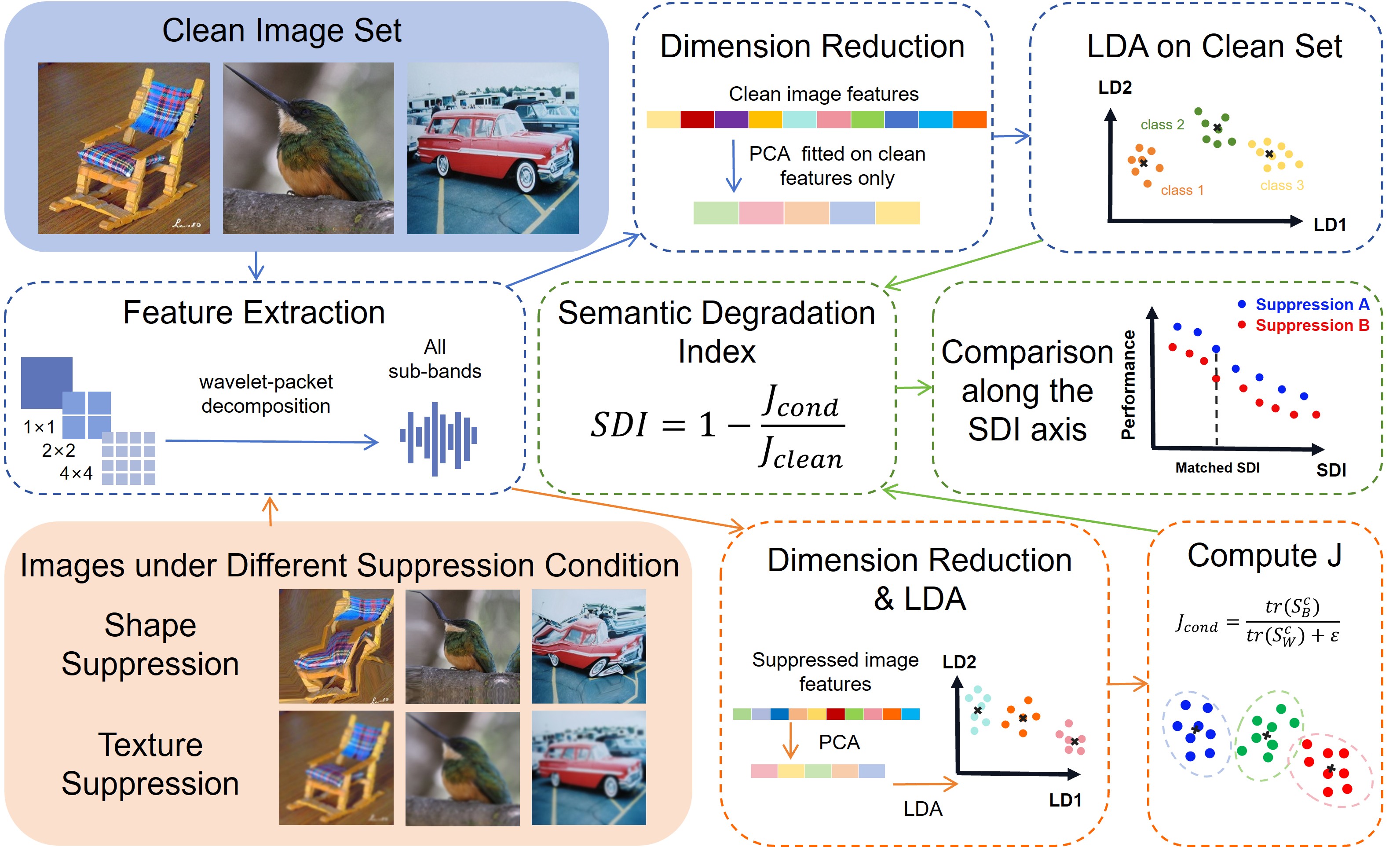}
    \caption{Overview of the Semantic Degradation Index (SDI) pipeline. (1) Clean images and their shape- and texture-suppressed counterparts are processed by the same spatial-pyramid wavelet-packet feature extractor. (2) Feature standardization, PCA, and LDA are fitted on the clean image set to define a fixed clean-reference discriminative space; suppressed images are projected using these same transformations. In the schematic LDA plots, colors denote categories, points denote image samples, crosses denote category centroids, and dashed ellipses illustrate within-class dispersion. Although two dimensions are shown for visualization, category separability is computed over the full LDA space from the ratio of between-class to within-class scatter. SDI measures the suppression-induced reduction in category separability relative to the clean reference. (3) Each Gaussian-blur or grid-distortion parameter combination yields an SDI value and a corresponding model-performance value. Shape- and texture-suppression trends are then compared over their overlapping SDI range.}
    \label{fig:wavelet+PCA+LDA}
\end{figure*}

\subsection{Comparison based on suppression-induced reductions in category separability}

Performance loss under suppression depends jointly on model reliance and category damage caused by the transformation. Because Gaussian blur and grid distortion are controlled by different parameter spaces, their nominal suppression strengths are not directly comparable. We therefore place all suppression conditions on a common axis defined by the loss of clean-reference category separability.

\begin{figure*}[t]
    \centering
    \includegraphics[width=0.85\textwidth]{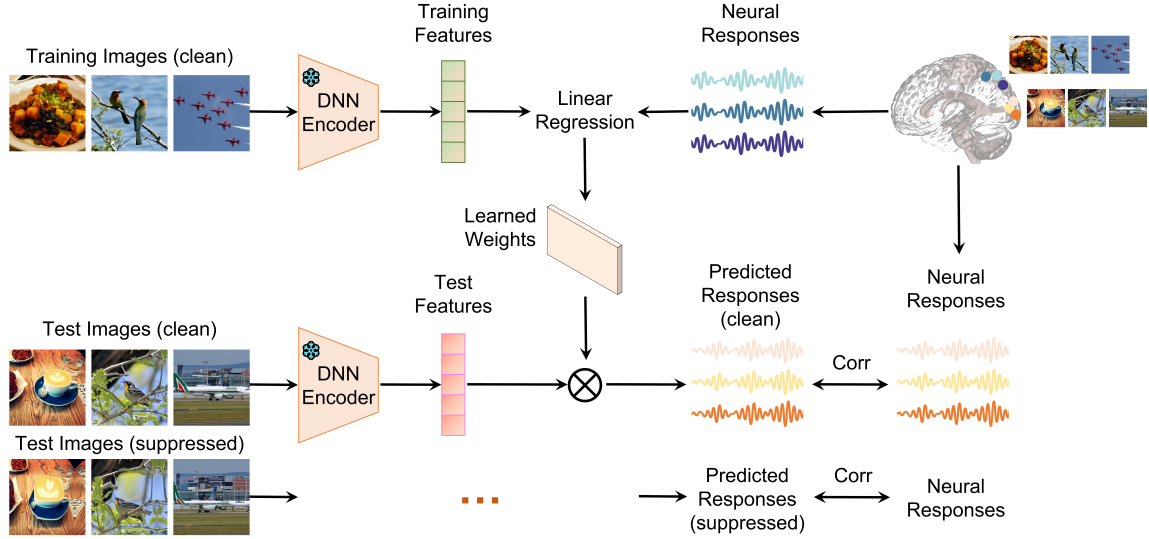}
    \caption{Brain-encoding evaluation under feature suppression. DNN features from clean training images are mapped to measured cortical responses using participant-specific vertex-wise linear encoding models. At test time, clean images and their suppressed counterparts are processed using the same DNN encoder and fixed encoding weights. Predictions from both conditions are evaluated against the measured responses to the original held-out images, and the suppression effect is quantified as the decrease from clean to suppressed encoding performance.}
    \label{fig:brain-encoding-method}
\end{figure*}

\textbf{(1) Clean-reference discriminative space.}  
As illustrated in Fig.~\ref{fig:wavelet+PCA+LDA}, clean and suppressed images are processed using the same handcrafted feature extractor. Each image is resized to $128 \times 128$ pixels using bilinear interpolation and represented in RGB. A spatial pyramid with $1 \times 1$, $2 \times 2$, and $4 \times 4$ partitions produces 21 spatial regions. Within each region, a three-level db2 wavelet-packet decomposition~\citep{mallat1989wavelet} is applied separately to each color channel using symmetric boundary extension. All terminal subbands are retained. Their coefficient magnitudes are transformed using $\log(1+x)$, flattened, and concatenated into a single feature vector.

The full clean benchmark is used to define the reference space and its baseline category separability. Feature-wise means and standard deviations are estimated from the clean features and applied to all suppression conditions. Principal component analysis (PCA)~\citep{jolliffe2002pca} is fitted on the standardized clean features, retaining up to 512 principal components. Linear discriminant analysis (LDA)~\citep{fisher1936measurements} is then fitted on the clean PCA features to define a discriminative space for the 16 categories. With 16 categories, the resulting space contains at most 15 discriminant dimensions. Clean and suppressed images are projected using the same clean-derived standardization, PCA, and LDA transformations.

\textbf{(2) Semantic Degradation Index.}
Let $\mathbf{z}_i^c$ denote the representation of image $i$ under condition $c$ in the fixed discriminative space. For category $k$, let $n_k$ denote the number of images, $\boldsymbol{\mu}_k^c$ the category mean, and $\boldsymbol{\mu}^c$ the mean across all images. The between-class and within-class scatter under condition $c$ are defined as:

\[
S_B^c
=
\sum_{k=1}^{K}
n_k
\left\|
\boldsymbol{\mu}_k^c-\boldsymbol{\mu}^c
\right\|_2^2,
\]

\[
S_W^c
=
\sum_{k=1}^{K}
\sum_{i:y_i=k}
\left\|
\mathbf{z}_i^c-\boldsymbol{\mu}_k^c
\right\|_2^2,
\]

Category separability is summarized by:

\[
J_{cond}
=
\frac{S_B^c}{S_W^c+\epsilon},
\]

where $\epsilon$ is a small constant for numerical stability. Let $J_{\mathrm{clean}}$ denote the corresponding value computed from the clean image set. The Semantic Degradation Index is defined as:

\[
\mathrm{SDI}_c
=
1-
\frac{J_\mathrm{cond}}{J_{\mathrm{clean}}}.
\]

An SDI of zero indicates unchanged category separability relative to the clean reference, whereas larger positive values indicate a greater suppression-induced reduction in category separability.

\textbf{(3) Comparison along the SDI axis.}
Each Gaussian-blur or grid-distortion parameter combination defines one suppression condition. For each condition, SDI and model performance are computed independently, producing one point in the SDI-performance plane. Performance trends are examined separately for shape and texture suppression and compared over their overlapping SDI range. SDI therefore provides a common horizontal axis for comparing model performance at similar estimated reductions in clean-reference category separability. Supplementary analyses based on Mahalanobis distance, centroid margin, and Bhattacharyya distance show trends broadly consistent with SDI. We therefore use SDI as the primary measure in the main analysis.

\subsection{Classification-based feature-reliance evaluation}

Model predictions are mapped from ImageNet subclasses to the corresponding 16 entry-level categories used in the ImageNet16-like benchmark. For each suppression condition, classification performance is reported as the percentage of accuracy retained relative to the model's clean-image accuracy.

Each Gaussian-blur or grid-distortion condition produces an SDI value and a corresponding retained-accuracy value. Shape- and texture-suppression trends are compared over their overlapping SDI range. At comparable SDI values, lower retained accuracy under one suppression type indicates stronger reliance on the suppressed information. CNNs and ViTs are compared within each suppression family, with retained accuracy serving as the measure of suppression robustness.

\subsection{Brain-encoding evaluation under feature suppression}

Brain-encoding analysis is performed using image-evoked fMRI responses from the eight NSD participants. As illustrated in Fig.~\ref{fig:brain-encoding-method}, DNN features extracted from clean training images are linearly mapped to cortical responses using participant- and vertex-specific encoding models~\citep{naselaris2011encoding}. The learned encoding weights remain fixed during evaluation.

Clean held-out images and their suppressed counterparts are passed through the same DNN encoder and encoding models. Both predictions are evaluated against the measured response to the corresponding original image. Encoding performance was quantified as noise-ceiling-normalized explained variance, following Gifford et al.~\citep{gifford20257t}. Pearson correlations between predicted and measured responses were computed across test images for each participant and cortical vertex. Negative correlations were set to zero, squared to obtain explained variance, and divided by the corresponding vertex-wise noise ceiling. The suppression effect is defined as the decrease from clean to suppressed encoding performance. CNNs and ViTs are compared separately under shape and texture suppression. Further details on feature extraction, encoding-model fitting, vertex selection, and aggregation are provided in the supplementary material.

\begin{table}[t]
\centering
\begin{tabular}{lcccc}
\toprule
\textbf{Condition}
& \multicolumn{2}{c}{\textbf{Texture}}
& \multicolumn{2}{c}{\textbf{Shape}} \\
\cmidrule(lr){2-3}
\cmidrule(lr){4-5}
& \textbf{LV} & \textbf{HFE} & \textbf{ESSIM} & \textbf{GC} \\
\midrule
Grayscale       & 0.995 & 0.994 & 0.996 & 0.995 \\
Gaussian blur   & 0.389 & 0.303 & 0.787 & 0.567 \\
Grid distortion & 0.811 & 0.817 & 0.223 & 0.015 \\
\bottomrule
\end{tabular}
\caption{Normalized low-level metrics under representative suppression conditions. The clean condition is normalized to $1$. Lower LV/HFE indicate stronger texture attenuation, lower ESSIM/GC indicate stronger shape disruption. (Settings: Gaussian blur, $k=11$ and $\sigma=2.0$; grid distortion, 8 grid steps and displacement limit $0.8$.)}
\label{tab:pilot_suppression_validation}
\end{table}

\section{Experiments and Results}

\subsection{CNNs show stronger texture than shape reliance at similar SDI values}

\paragraph{Pilot study: operator validation and grayscale influence.}
Following prior work~\citep{burgert2026imagenet}, we evaluated representative grayscale, Gaussian-blur, and grid-distortion conditions using LV and HFE as texture-related metrics and ESSIM and GC as shape-related metrics.

As shown in Table~\ref{tab:pilot_suppression_validation}, Gaussian blur most strongly reduced LV and HFE, whereas grid distortion most strongly reduced ESSIM and GC. Grayscale conversion left all four metrics near the clean baseline. These results support the use of Gaussian blur and grid distortion for texture and shape suppression, respectively.
Using the same representative conditions, we evaluated retained accuracy across ten ImageNet-trained models. Grayscale conversion caused the smallest performance decrease for every model, with the same pattern observed for both CNNs and ViTs (Table~\ref{tab:operators_models}).
The subsequent matched analyses therefore focus on shape and texture suppression.

\begin{table}[t]
\centering
\begin{tabular}{lccc}
\toprule
\textbf{Model}
& \textbf{Blur}
& \textbf{Distortion}
& \textbf{Grayscale} \\
\midrule
ResNet50                & 64.13 & 72.27 & 87.57 \\
ConvNeXt                & 64.42 & 74.52 & 89.84 \\
EfficientNet            & 75.49 & 72.65 & 94.12 \\
InceptionV3             & 66.51 & 70.00 & 90.43 \\
MobileNetV4             & 70.96 & 72.09 & 95.42 \\
Swin-B                   & 79.51 & 84.21 & 96.91 \\
BEiTv2                   & 86.63 & 87.17 & 98.78 \\
DeiT3                    & 93.47 & 85.27 & 99.13 \\
ViT-Imagenet             & 73.07 & 78.36 & 90.27 \\
ViT-CLIP                 & 85.19 & 85.91 & 95.14 \\
\midrule
CNN mean                 & 68.30 & 72.31 & 91.48 \\
ViT mean                 & 83.57 & 84.18 & 96.05 \\
\bottomrule
\end{tabular}
\caption{Retained classification accuracy (\%) under representative suppression conditions.}
\label{tab:operators_models}
\end{table}

\paragraph{Experimental setup.} Five ImageNet-trained CNNs (ResNet50~\cite{he2016deep}, ConvNeXt~\cite{liu2022convnet}, EfficientNet~\cite{tan2021efficientnetv2}, InceptionV3~\cite{szegedy2016rethinking}, and MobileNetV4~\cite{qin2024mobilenetv4}) were evaluated across the Gaussian-blur and grid-distortion parameter grids. Each suppression condition produced one SDI-retained-accuracy point, and the two suppression families were compared over their overlapping SDI range.

\begin{figure}[t]
    \centering
    \includegraphics[width=\linewidth]{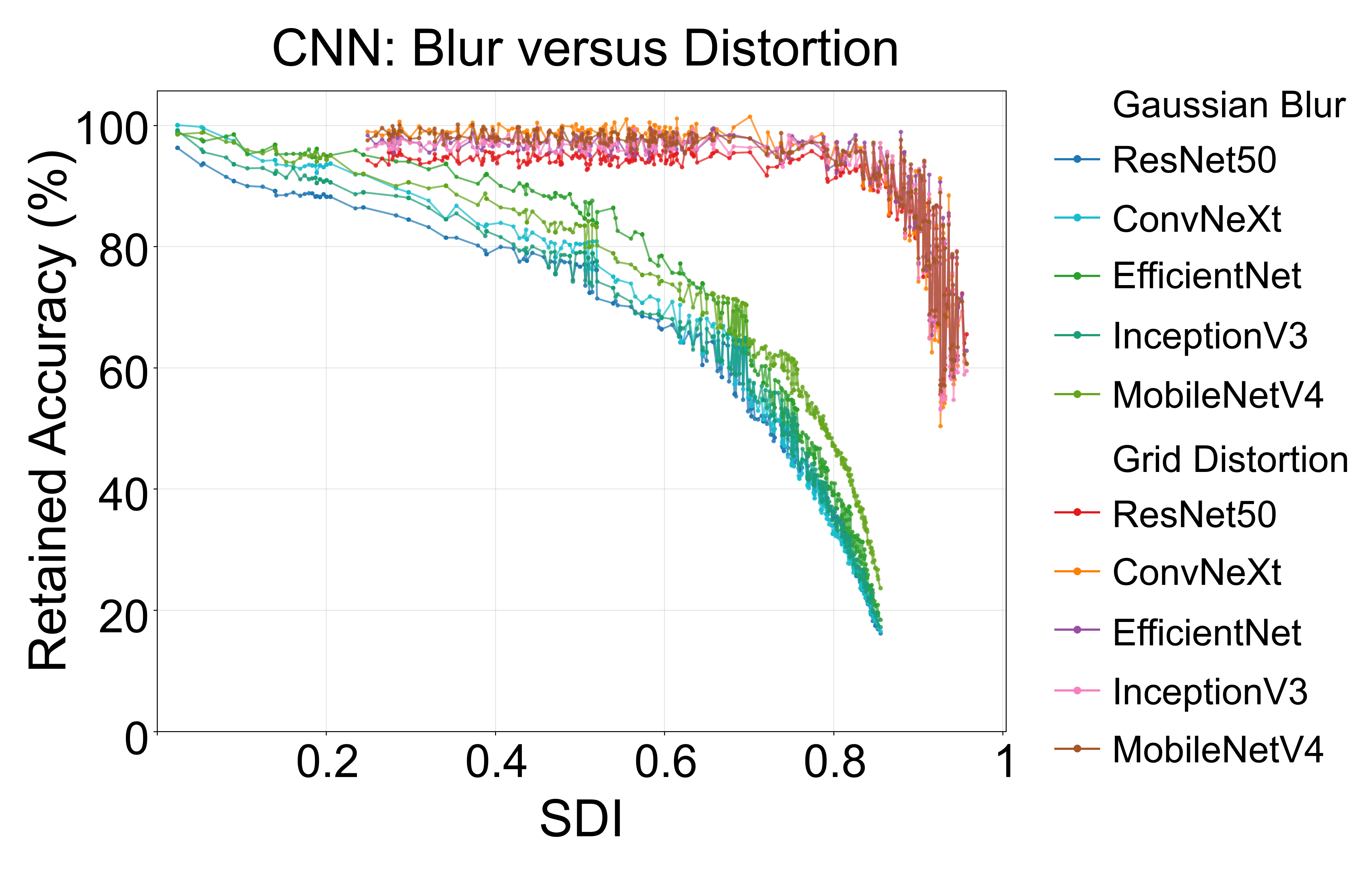}
    \caption{Retained accuracy of five CNNs under Gaussian blur and grid distortion. Gaussian blur yields lower accuracy over the overlapping SDI range, indicating greater sensitivity to texture suppression.}
    \label{fig:Experiments-cnn-1}
\end{figure}

\paragraph{Main results.} As shown in Fig.~\ref{fig:Experiments-cnn-1}, Gaussian blur produced consistently lower retained accuracy than grid distortion for all five CNNs at comparable SDI values. After accounting for the estimated loss of clean-reference category structure, CNN classification was therefore more sensitive to texture suppression than to shape suppression. This finding extends prior suppression-based analyses by showing that the relative sensitivity to shape and texture suppression can change when the two are compared at similar levels of category damage.

\subsection{ViTs retain higher accuracy under shape and texture suppression}

\begin{figure*}[t]
    \centering
    \includegraphics[width=0.8\linewidth]{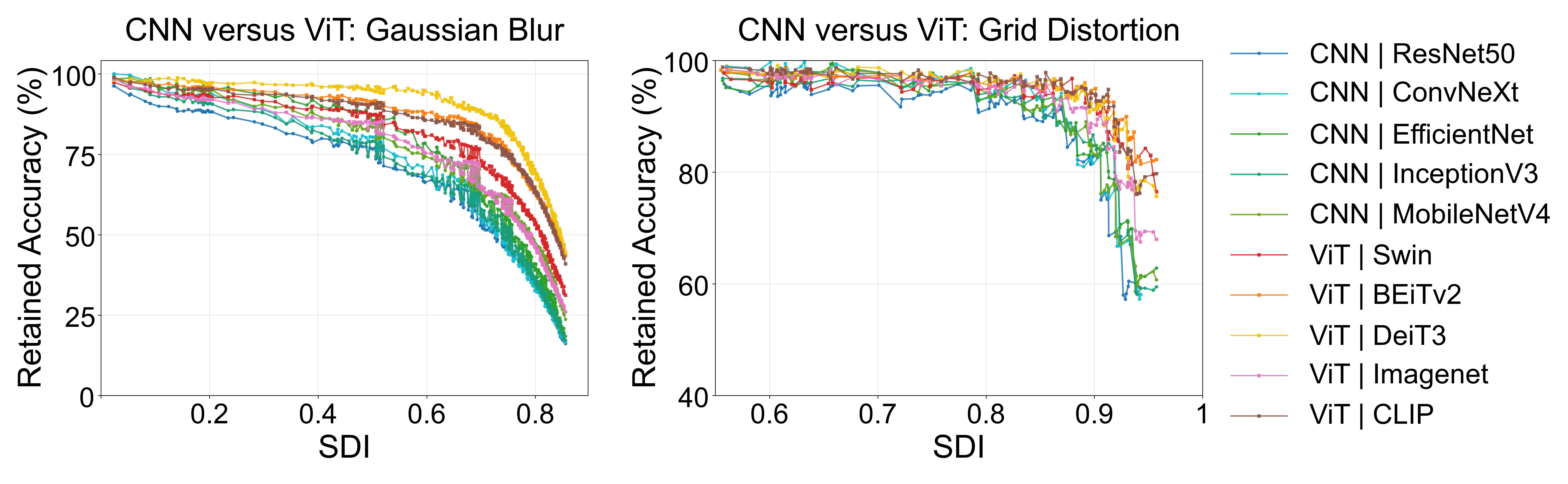}
    \caption{Retained accuracy of five CNNs and five ViTs as a function of SDI. ViTs retain higher accuracy across most of the Gaussian-blur range and show a clearer advantage under grid distortion at high SDI.}
    \label{fig:cnn-vit-sdi}
\end{figure*}

We compare five ImageNet-trained CNNs and five ViTs (Swin-B~\cite{liu2021swin}, BEiTv2~\cite{peng2022beit}, DeiT3~\cite{touvron2022deit}, ViT-Imagenet~\cite{dosovitskiy2020image} and ViT-CLIP~\cite{radford2021learning}) by expressing retained accuracy as a function of SDI for each suppression family. Under Gaussian blur, ViTs retain higher accuracy than CNNs across most of the available SDI range, with the difference increasing as category separability decreases (Fig.~\ref{fig:cnn-vit-sdi}). Under grid distortion, the two model families show similar performance at lower SDI values, whereas CNN accuracy declines more sharply at high SDI and most ViTs retain higher accuracy (Fig.~\ref{fig:cnn-vit-sdi}). 

\subsection{ViTs show smaller suppression-induced declines in brain encoding}

Using the representative Gaussian-blur and grid-distortion conditions from the pilot study, we compared suppression-induced changes in noise-ceiling-normalized explained variance for five CNNs and five ViTs. Encoding maps were averaged across the eight NSD participants and then across models within each family. White contours indicate visual regions from the NSD streams ROI collection; the labeled parcellation is provided in the supplementary material.

\begin{figure}[t]
    \centering
    \includegraphics[width=\linewidth]{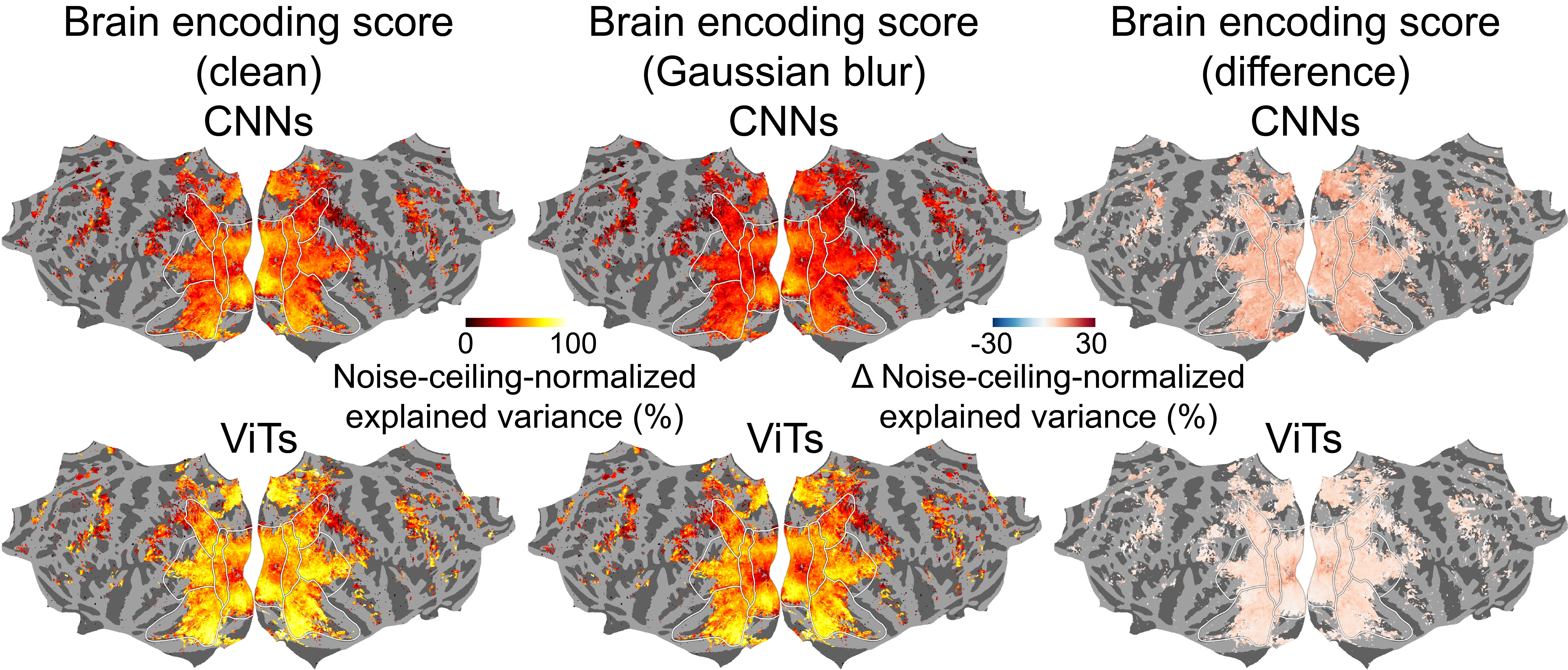}
    \caption{Group-averaged brain-encoding results under Gaussian blur. Rows show CNNs and ViTs; columns show noise-ceiling-normalized explained variance for clean and blurred images and their clean-minus-blurred difference. Positive differences indicate reduced encoding performance after suppression. White contours mark visual regions from the NSD streams ROI collection.}
    \label{fig:brain-fig-gau-group}
\end{figure}

\begin{figure}[t]
    \centering
    \includegraphics[width=\linewidth]{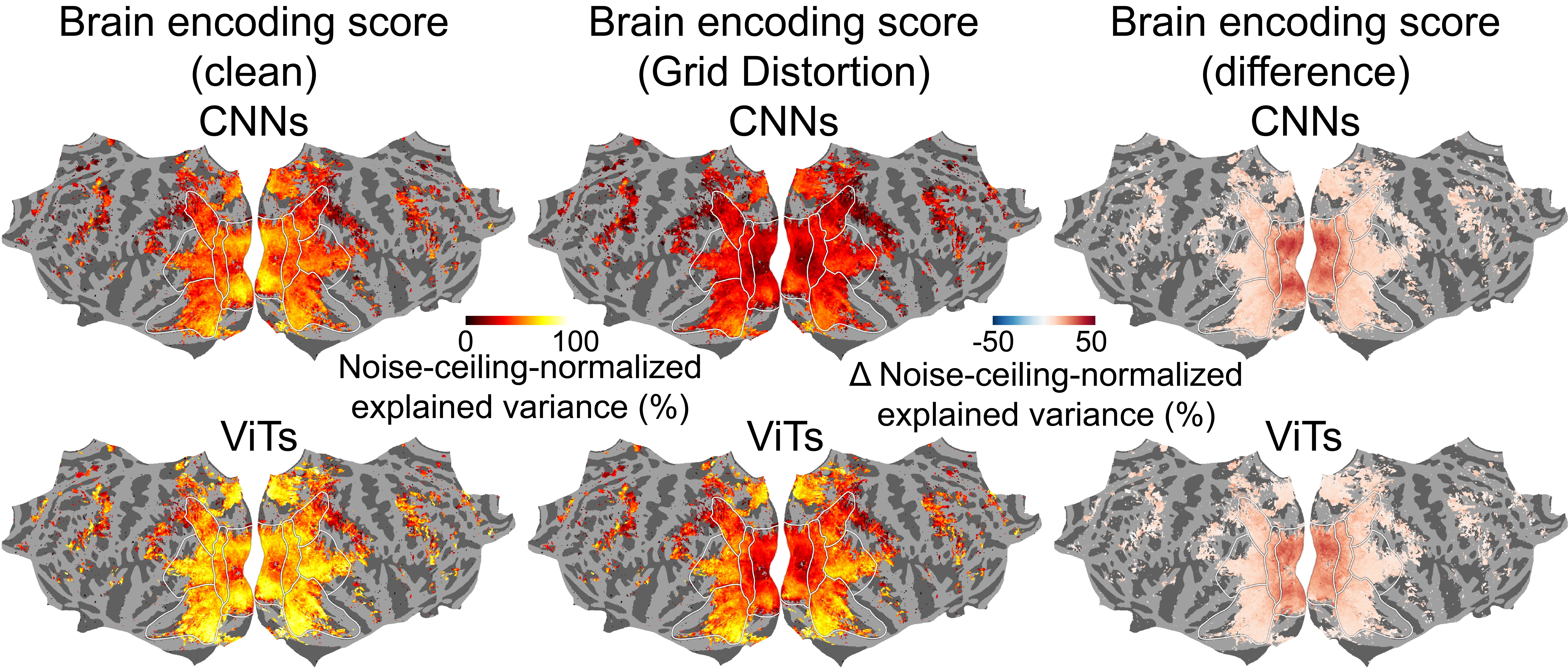}
    \caption{Group-averaged brain-encoding results under grid distortion. Rows show CNNs and ViTs; columns show noise-ceiling-normalized explained variance for clean and distorted images and their clean-minus-distorted difference. Positive differences indicate reduced encoding performance after suppression. White contours mark visual regions from the NSD streams ROI collection.}
    \label{fig:brain-fig-shape-group}
\end{figure}

Gaussian blur reduced encoding performance for both model families, but the clean-minus-blurred differences were smaller for ViTs than for CNNs (Fig.~\ref{fig:brain-fig-gau-group}). Grid distortion produced the same model-family pattern, with larger decreases for CNNs across the outlined visual regions (Fig.~\ref{fig:brain-fig-shape-group}). Thus, the higher retained classification accuracy of ViTs was accompanied by smaller suppression-induced declines in Brain encoding. Model-wise maps are provided in the supplementary material.

\section{Conclusion}



Feature-suppression studies infer reliance from the performance loss caused by attenuating selected visual information. Comparisons across feature types, however, are difficult to interpret when their suppression operators reduce category-relevant information by different amounts. We introduced the Semantic Degradation Index (SDI) to quantify this loss relative to clean images in a fixed clean-reference discriminative space, thereby providing a shared category-level axis for comparing otherwise different transformations.

At comparable SDI values, all five evaluated ImageNet-trained CNNs retained less accuracy under Gaussian blur than under grid distortion. Under the evaluated operators, this pattern supports stronger texture than shape reliance and contrasts with the shape-dominant conclusion obtained from unmatched suppression conditions~\citep{burgert2026imagenet}. The evaluated ViTs generally retained more accuracy than CNNs, with an advantage across most of the Gaussian-blur range and a clearer advantage under grid distortion at high SDI. Under representative conditions for both operators, ViT features also showed smaller decreases in noise-ceiling-normalized explained variance. The CNN--ViT difference under feature suppression therefore extends beyond classification outputs to representations predictive of human visual cortical responses.

These findings motivate separating two questions in suppression-based evaluation: whether an operator changes its intended visual property, and how much category structure it removes while doing so. Feature-specific metrics address the first question, whereas SDI provides a common reference for the second. SDI reduces an important cross-operator confound, but it does not make the transformations identical or guarantee perfect isolation of shape and texture; the conclusions remain conditional on the evaluated operators, models, dataset, and reference feature space. More broadly, calibrating interventions by their category-level consequences provides a more controlled basis for interpreting feature reliance and comparing model robustness.

\bibliography{reference}

\end{document}